\documentclass[letterpaper]{article} 
\usepackage[]{aaai2027_arxiv}  
\usepackage[hyphens]{url}  
\usepackage{graphicx} 
\usepackage{natbib}  
\usepackage{caption} 
\usepackage{amsmath}
\usepackage{amssymb}
\usepackage{bm}
\usepackage{booktabs}
\usepackage{multirow}
\usepackage{subcaption}
\usepackage{xcolor}
\usepackage{algorithm}
\usepackage{algorithmic}

\title{Quality Action Assurance: Multimodal Verification of Examiner Claims in VR OSCEs}

\author{
    Harry Rogers\textsuperscript{\rm 1},
    Sally Shiels\textsuperscript{\rm 2},
    Ashley Tomlinson\textsuperscript{\rm 2},
    James Thomas\textsuperscript{\rm 2},
    James Aylward\textsuperscript{\rm 3},\\
    Nathan Gauge\textsuperscript{\rm 2},
    Helen Higham\textsuperscript{\rm 2},
    Alison Noble\textsuperscript{\rm 1}
}
\affiliations{
    \textsuperscript{\rm 1}Department of Engineering Science, University of Oxford, Oxford, UK\\
    \textsuperscript{\rm 2}Nuffield Department of Clinical Neurosciences, University of Oxford, Oxford, UK\\
    \textsuperscript{\rm 3}Department for Continuing Education, University of Oxford, Oxford, UK\\
    \{harry.rogers, alison.noble\}@eng.ox.ac.uk\\
    \{sally.shiels, ashley.tomlinson, james.thomas, nathan.gauge, helen.higham\}@ndcn.ox.ac.uk\\
    james.aylward@conted.ox.ac.uk
}

\begin{document}

\maketitle

\begin{abstract}
Objective Structured Clinical Examinations (OSCEs) are the gold standard for assessing clinical competence, yet scoring remains vulnerable to examiner subjectivity, fatigue, and cognitive bias. Standard examiner validation via inter-rater statistics lacks explanatory power regarding the source of errors, as it neither analyzes examiner reasoning nor verifies examiner claims against actual events. Thus, we introduce Quality Action Assurance (QAA), a multimodal framework that verifies examiner claims in Virtual Reality (VR) pediatric OSCEs by comparing actions claimed by examiners against a reference record of events constructed from video, VR logs, and actor annotations. QAA combines a constrained temporal action alignment model, which performs action localization and actor source attribution, with a large language model that extracts examiner claims and checks them against the record. Across a 5-fold cross-validation, QAA achieves 99.2\% $\pm$ 0.7\% Actor F1 and 93.4\% $\pm$ 1.9\% W@16 for temporal alignment. Overall, QAA detects examiner errors with 69.9\% precision and 76.7\% recall; in retrospective evaluation, correcting the detected errors raises the share of factually correct transcripts from 39.2\% to 79.2\%, supporting fairer OSCE quality assessment.
\end{abstract}

\section{Introduction}
Objective Structured Clinical Examinations (OSCEs) are the gold standard for certifying clinical competence \cite{khan2013osce}. Virtual Reality (VR) OSCEs offer scalable standardization through consistent scenario playback \cite{kelly2025vr,stehling2025vr}; yet, assessment still depends on human examiners observing, interpreting, and judging under time pressure. Marking is susceptible to cognitive noise \cite{castor2024measuring,alnabelsi2024inter}: examiners may reconstruct events using schemas rather than faithfully recording them \cite{tversky1974judgment,reason1990human}, producing systematic factual errors undetectable by inter-rater reliability.

Most published work on automated assessment in Surgical Data Science and OSCE Artificial Intelligence (AI) targets \emph{student} performance from video, kinematics, or logs \cite{maier2017surgical,abou2025egocentric,ma2025ai,ecosbot2024}, implicitly treating expert judgments as a ground truth despite known label variability \cite{maierhein2022sds}. Psychometric studies monitor extreme examiner effects \cite{fuller_osce_extremes}, while cognition-focused work uses think-aloud protocols to characterize scoring assumptions \cite{chahine2016_minds,rodutaRoberts2020_assessorCognition,scully2022_thesis}. These approaches document that examiner cognition contributes to variance but do not verify whether specific claimed reasoning is factually correct. Closer work interrogates expert decisions directly \cite{alur2024auditinghumanexpertise,Mullainathan2021-hv}, but assesses decision quality in aggregate; we verify the individual factual claims an examiner makes about one performance.

Automating the assessment itself is not the answer: OSCE assessment is fundamentally a complex observation task extending beyond a checklist of actions. Examiners must track the sequence of events, the specific actors addressed, and qualitative aspects of performance: a student's confidence, decisiveness, and timeliness are clinically meaningful proxies for safety, and a technically correct action performed with hesitation may still indicate poor practice. System logs, such as ones within VR, record \emph{that} an action occurred, not \emph{how} it was performed, so they cannot replace examiner judgment. Examiners must therefore remain the assessors, yet every qualitative judgment rests on factual premises about what was done, when, and by whom, and it is precisely these premises that can be checked against a logged event record. To understand how often those premises fail and how failures affect grades, we combined psychometrics with grounded verification: we examined inter-rater reliability alongside examiner verbal reasoning, and checked verbalized claims against an actor-attributed event record. Using concurrent and retrospective verbal protocols \cite{ericsson1993protocol}, we identified two recurring cognitive errors: (i) Inferential, where an examiner asserts an action occurred based on suggestive cues rather than definitive completion \cite{schacter1999sins}, and (ii) Source Misattribution, where the examiner identifies the correct action but assigns it to the wrong actor \cite{Johnson1997-np}.
\begin{figure*}[t]
\centering
\begin{subfigure}[b]{0.3\textwidth}
    \includegraphics[width=\textwidth]{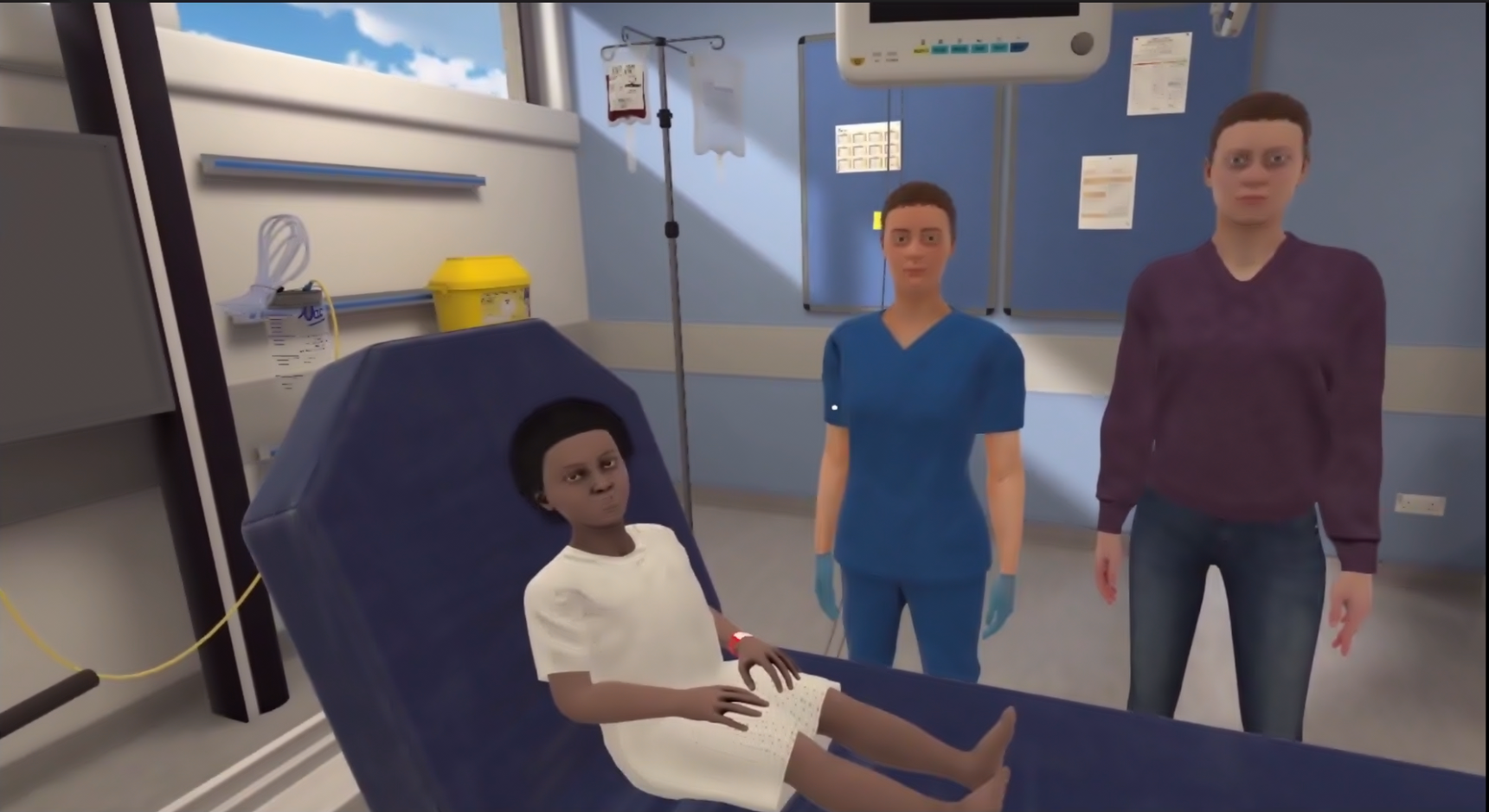}
    \caption{}
    \label{fig:fig1a}
\end{subfigure}
\hfill
\begin{subfigure}[b]{0.3\textwidth}
    \includegraphics[width=\textwidth]{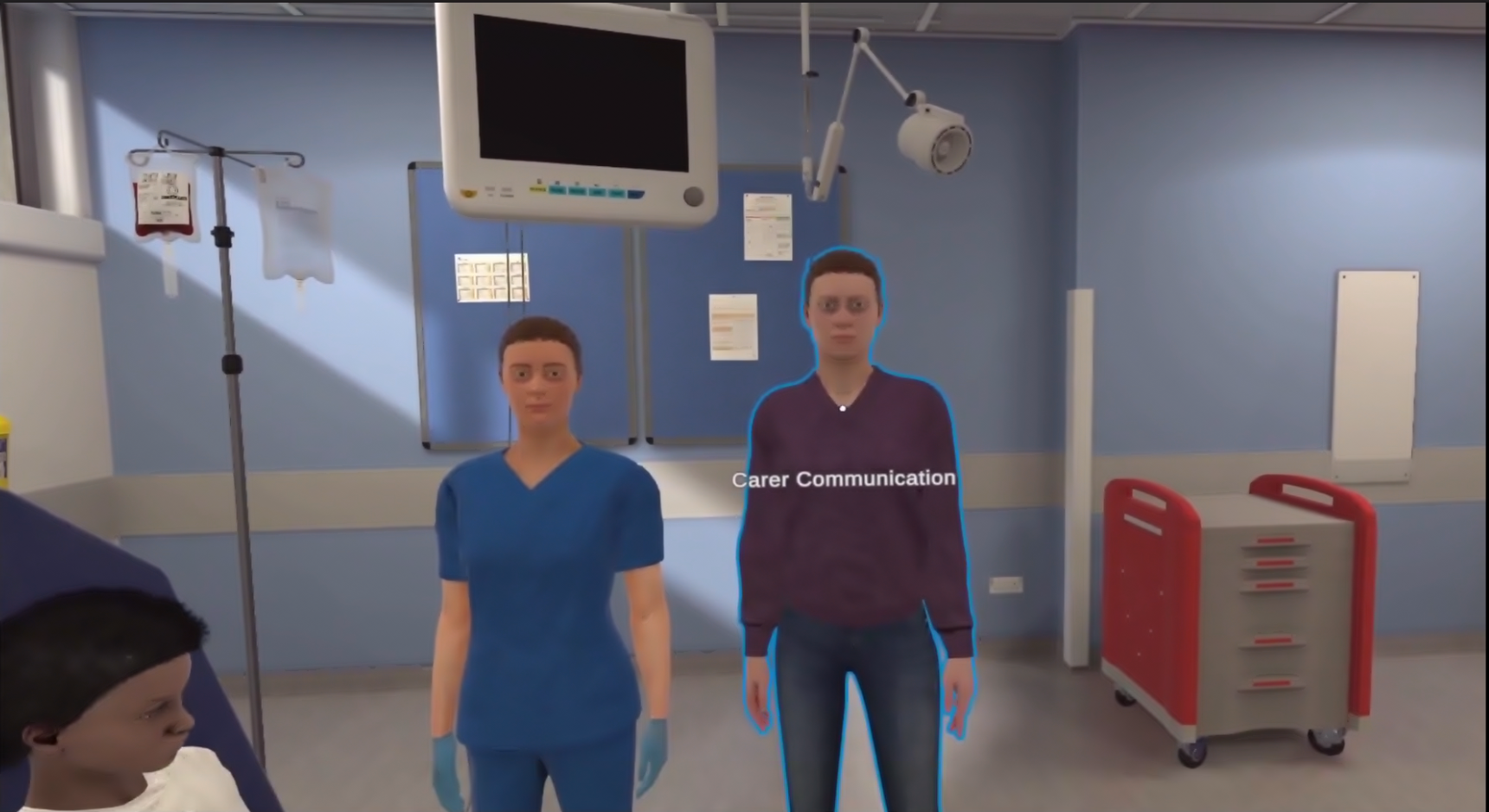}
    \caption{}
    \label{fig:fig1b}
\end{subfigure}
\hfill
\begin{subfigure}[b]{0.3\textwidth}
    \includegraphics[width=\textwidth]{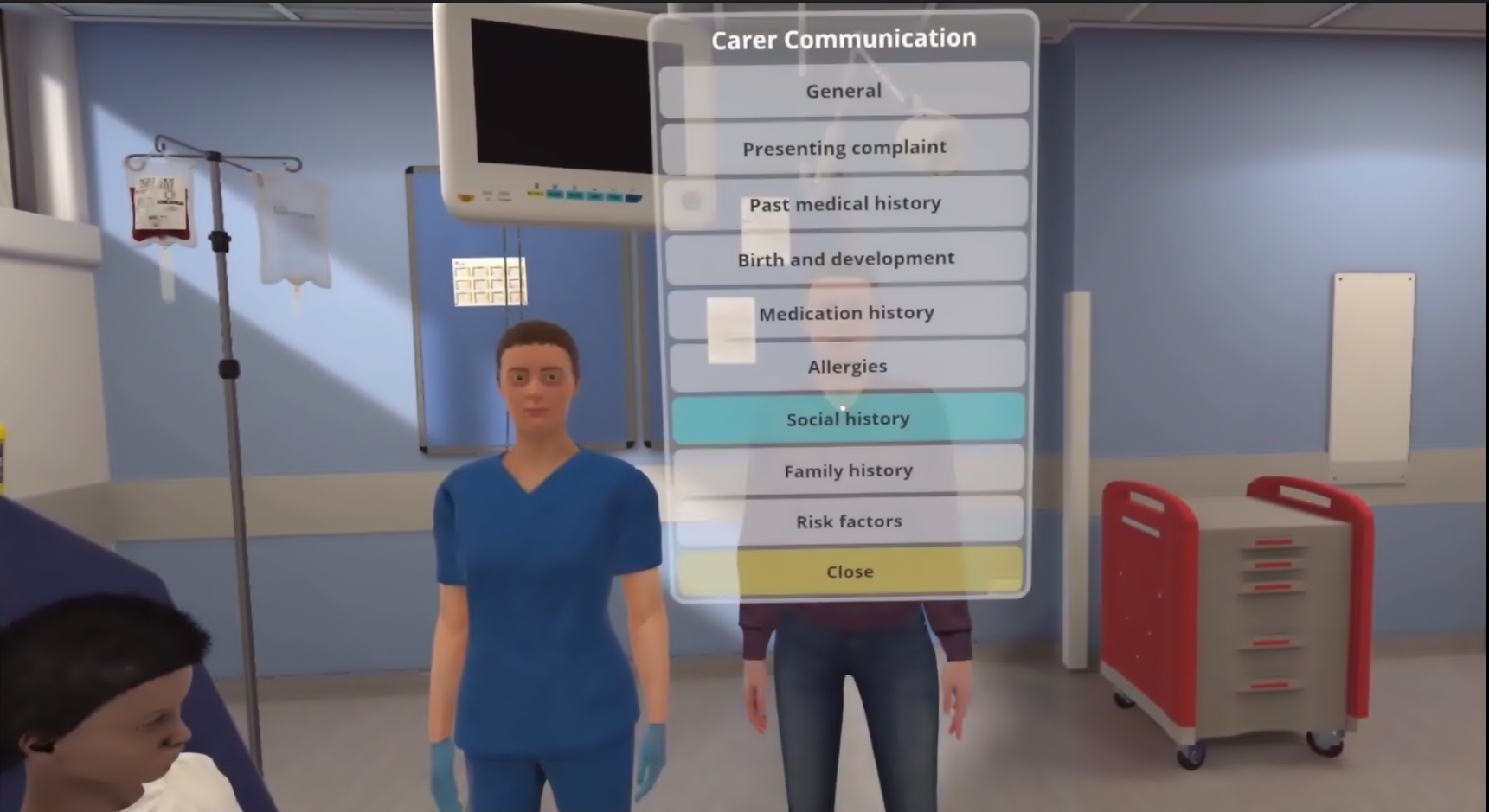}
    \caption{}
    \label{fig:fig1c}
\end{subfigure}
\caption{Egocentric view: (a) actors, (b) hover options, (c) action menu.}
\label{fig:fig1}
\end{figure*}
We introduce Quality Action Assurance (QAA), a multimodal framework that treats examiner verbalizations as verifiable claims and grounds them against an actor-attributed record derived from egocentric video aligned to VR action traces. QAA is designed as a decision-support system: it flags and explains potentially incorrect claims for human review rather than replacing examiner judgment. Our contributions are:
\begin{enumerate}
    \item A cognitive-error analysis of examiner reasoning in VR OSCEs, based on two-phase verbal protocols from four clinical examiners, revealing a \emph{reliability paradox}: examiners reach substantial-to-near-perfect agreement while making factual errors in over 60\% of assessments, errors large enough to shift the borderline-regression pass mark.
    \item A constrained temporal action alignment model that combines dual encoders, a monotonic alignment objective with a minimum temporal gap, and scenario knowledge constraints to localize each action and attribute it to the correct actor.
    \item An LLM verifier that extracts examiner claims from transcripts, checks them against the aligned actor-attributed record, and classifies mismatches as Inferential or Source Misattribution errors with an explicit rationale.
    \item A grading-impact analysis showing that correcting detected errors significantly changes per-student grades (exact sign test $p<10^{-3}$), lowers the cohort cut-score, and moves a borderline student from an automatic fail into the mandatory-adjudication zone.
\end{enumerate}

\section{Dataset and Cognitive Error Analysis}
\label{sec:data_acquisition}
We compiled egocentric videos and VR logs from 91 final year medical students completing a 12-minute pediatric emergency scenario on the OMS VR platform (Oxford Medical Simulation Ltd.). The study began before OMS introduced AI-first voice interaction; students therefore interacted with objects and three actors (Child, Carer, Nurse) via menus (Figure~\ref{fig:fig1}). This modality was methodologically useful because it generated visible, binary clicks, enabling examiner claims to be verified against the VR log. VR logs record action names but not target actors. For example, asking about allergies is logged identically whether directed to the child or carer, although only the carer reports the child's penicillin allergy. Thus, we manually relabeled all videos with actor labels (Child, Carer, Nurse, Item) and synchronized timestamps, producing an annotated reference record. Throughout, we distinguish the VR log, this annotated reference record, and the predicted record output by the alignment model. The Joint Research Office classified this as evaluation of educational provision, not requiring further ethical approval. Students and examiners consented; data are anonymized and simulation actors are automated avatars with pre-scripted behaviors.

\subsection{Verbal Protocols and Error Annotation}
\label{sec:expert_protocols}
Four clinical examiners each independently graded a subset of 30 anonymized students using a two-phase verbal protocol: (1) Concurrent observation, where examiners verbalized what they believed the student was doing and how it related to performance, and (2) Retrospective rubric completion, where examiners completed a multi-dimensional rubric covering eight assessment domains (History taking, Communication, Escalation, Physical Exam, Prescribing, Investigations, Management, Patient safety) and a Global outcome to retrospectively justify grades. Audio from both phases is segmented, transcribed with WhisperX \cite{bain2023whisperx}, and human-validated against what the examiner states.
\begin{table}[t]
\centering
\setlength{\tabcolsep}{3pt}
\begin{tabular}{l|r|r|r}
\toprule
\textbf{Examiner} & \textbf{Inferential} & \textbf{Source Misattribution} & \textbf{Total} \\
\midrule
A  & 21 & 8  & 29 \\
B  & 4  & 19 & 23 \\
C  & 14 & 13 & 27 \\
D  & 13 & 12 & 25 \\
\midrule
\textbf{Total} & \textbf{52} & \textbf{52} & \textbf{104} \\
\bottomrule
\end{tabular}
\caption{Examiner-level distribution of cognitive errors across assessments.}
\label{tab:error_analysis_experts}
\end{table}

\begin{figure*}[t]
\centering
\begin{subfigure}[b]{0.3\textwidth}
    \includegraphics[width=\textwidth]{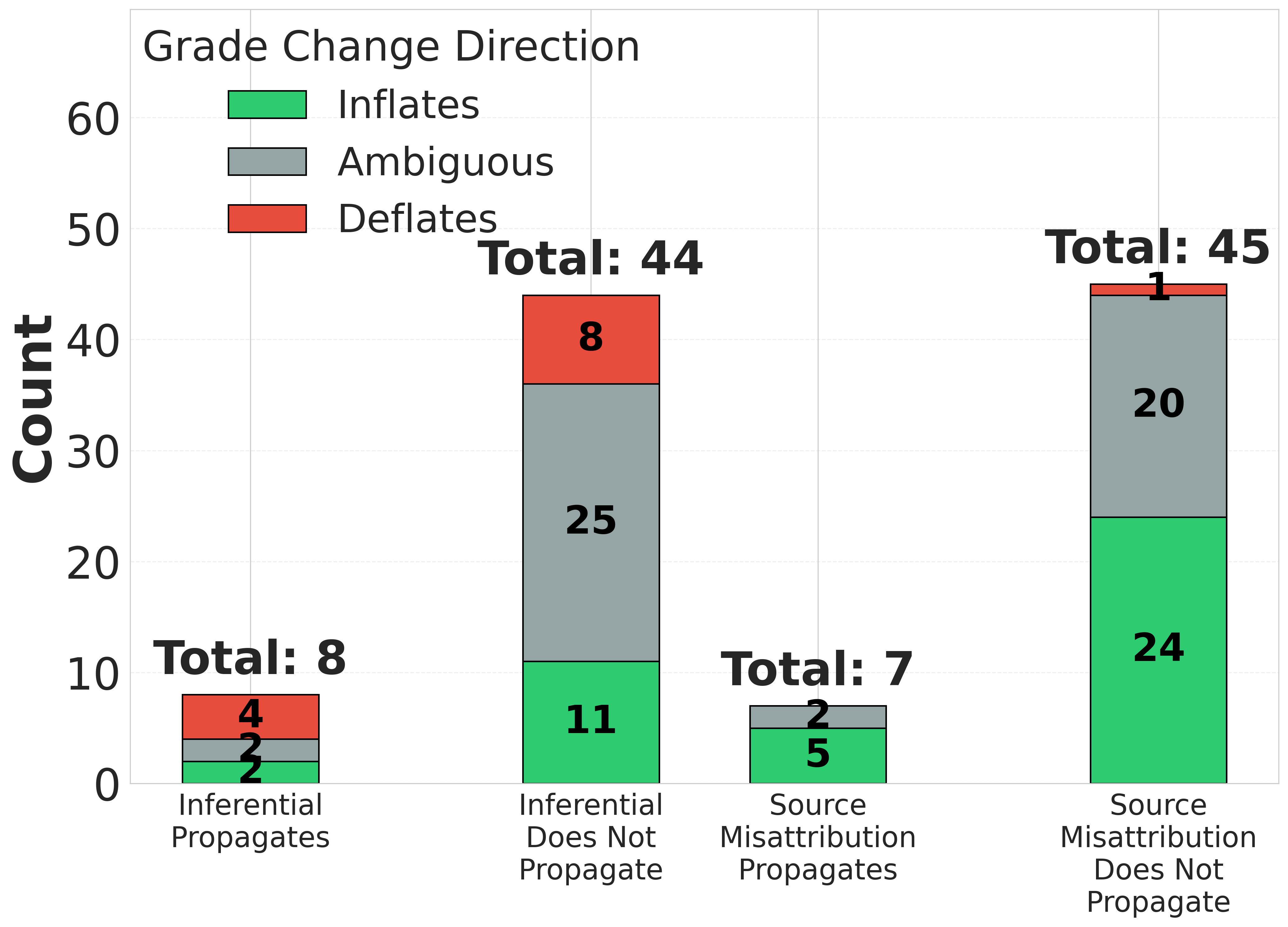}
    \caption{}
    \label{fig:fig2a}
\end{subfigure}
\hfill
\begin{subfigure}[b]{0.3\textwidth}
    \includegraphics[width=\textwidth]{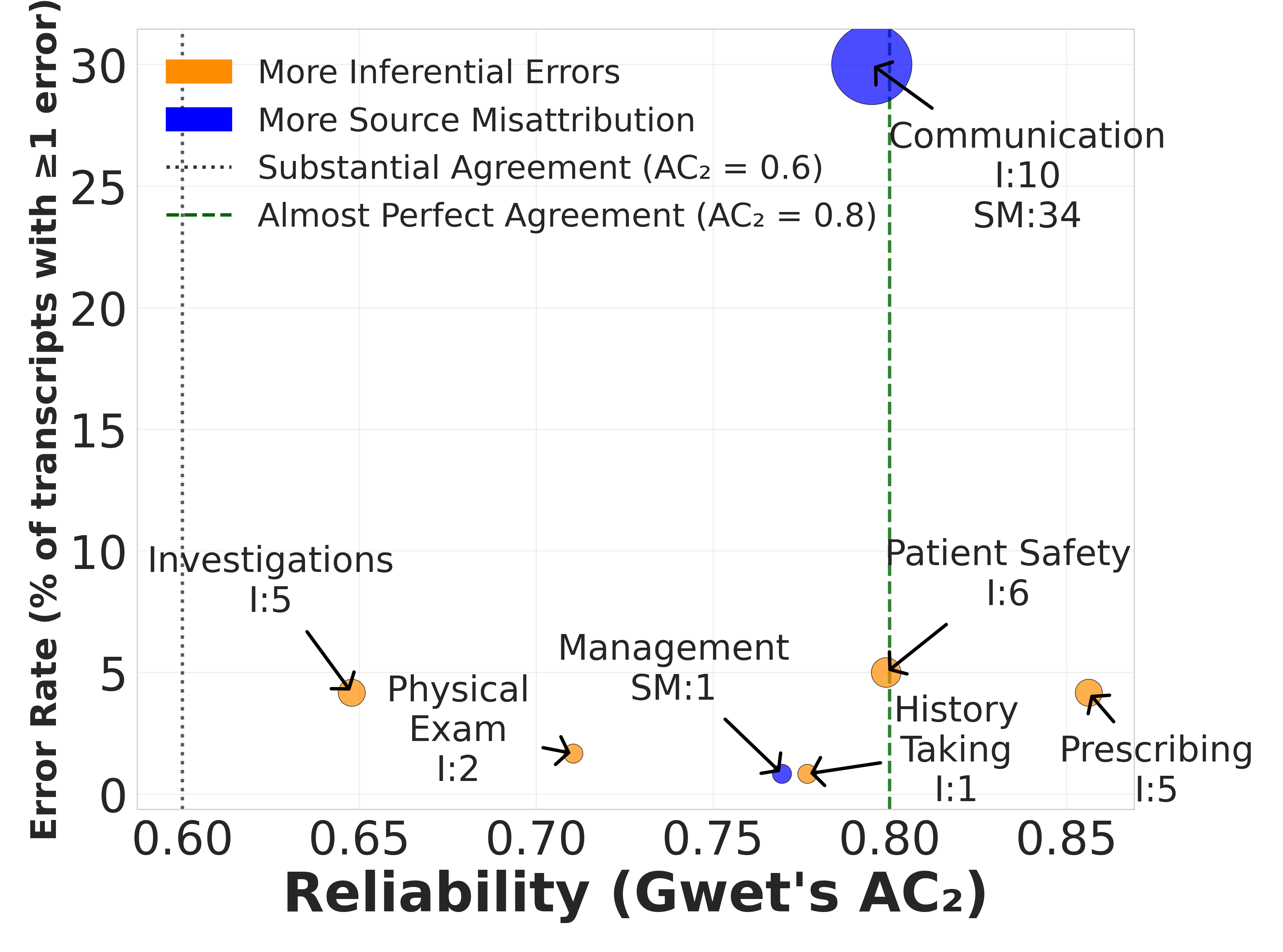}
    \caption{}
    \label{fig:fig2b}
\end{subfigure}
\begin{subfigure}[b]{0.3\textwidth}
    \includegraphics[width=\textwidth]{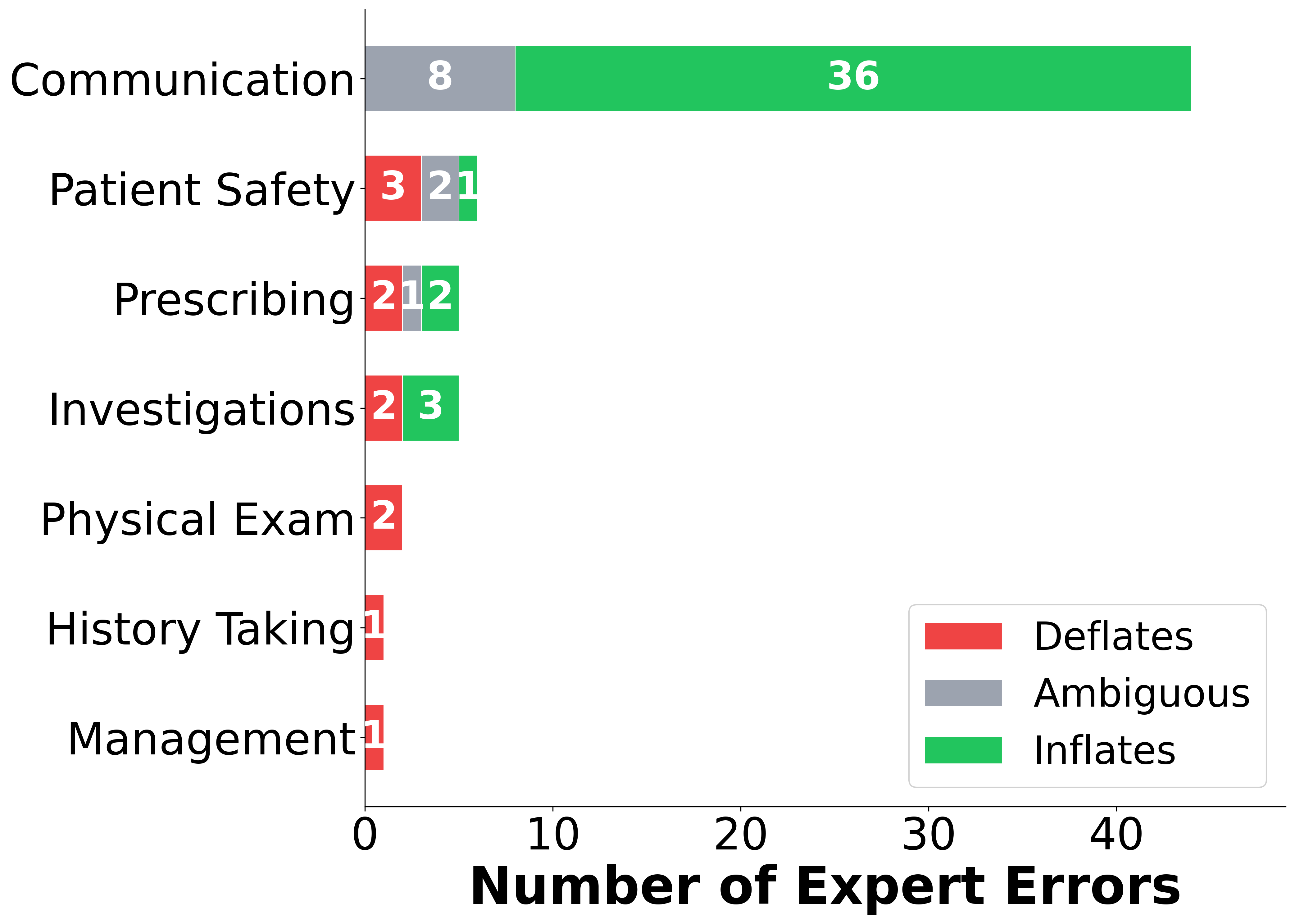}
    \caption{}
    \label{fig:fig2c}
\end{subfigure}
\caption{(a) Propagated and non-propagated errors with grading effect, (b) Domain error rate vs. inter-rater reliability, (c) Domain error counts with grading effect.}
\label{fig:fig2}
\end{figure*}
\paragraph{Taxonomy and Prevalence of Cognitive Errors.}
We grounded examiner claims against the annotated reference record to identify cognitive errors, defined as explicit claims unsupported by that record. We categorized each error as: (i) Inferential, where the examiner asserts an action occurred although it is absent from the record; or (ii) Source Misattribution, where the examiner correctly identifies the action but assigns it to the wrong actor, such as attributing a carer's statement to the child. Across the 120 transcripts, 73 (60.8\%) contained at least one error, totaling 104 errors shown in Table~\ref{tab:error_analysis_experts}; multiple errors often occurred within the same performance or recurred across different examiners assessing the same student. We further labeled errors across three dimensions to understand their impact: (1) Domain Mapping, identifying the clinical domain the examiner was discussing; (2) Propagation, labeling an error as propagated when the same claim appeared in both concurrent observation and retrospective rubric completion, indicating that an initial perceptual error persisted to directly influence the final assessment; and (3) Direction of Impact, analyzing the examiner's evaluative language to determine if the error inflated or deflated the student's score. For example, a propagated error resulting in grade deflation occurred where an examiner stated during observation, \textit{``They requested a lumbar puncture,''} and later penalized the student during rubric completion \textit{``...getting a lumbar puncture [was not] indicated... putting all of those at poor''} despite the student never requesting the procedure in the VR log.

\paragraph{The Reliability Paradox.}
We measured inter-rater reliability using Gwet's AC$_2$ with quadratic weights \cite{gwet2001handbook} to accommodate the non-linear ordinal scales: domain-specific grades are Poor, Satisfactory, Excellent, and the Global outcome is Clear Fail, Borderline Fail, Borderline Pass, Clear Pass, Above Expected. Despite substantial-to-near-perfect agreement (AC$_2$ $\approx$ 0.65--0.86), cognitive errors persist, creating a reliability paradox where examiners are statistically consistent yet factually incorrect, the gap validity theory draws between reliability and valid score interpretation \cite{messick1995validity,kane,downing}. These errors extend beyond commentary: of 104 total errors, 53 were grade-affecting across all domains except Escalation and the Global outcome. However, since domain scores serve as anchors for standard setting using borderline regression for OSCEs and surgical assessment \cite{Kramer2003-ve,De_Montbrun2015-wx}, these factual errors directly compromise the calculated pass mark. Applying an oracle correction, resolving all 53 grade-affecting errors, shifted the composite cut-score from 46.3\% to 45.2\% ($R^2 = 0.69$) and moved one student from \textbf{fail to pass}. Under this annotation-based counterfactual, the student's recorded failure is an artifact of examiner error rather than a lack of clinical ability. Crucially, even non-propagated errors influenced grades, suggesting examiners introduce factual errors from memory without prior verbalization. Figure~\ref{fig:fig2} visualizes the relationship between error rates, reliability, and grading effects.

\section{Methodology: Quality Action Assurance (QAA)}
\begin{figure}[t]
    \centering
    \includegraphics[width=0.99\columnwidth]{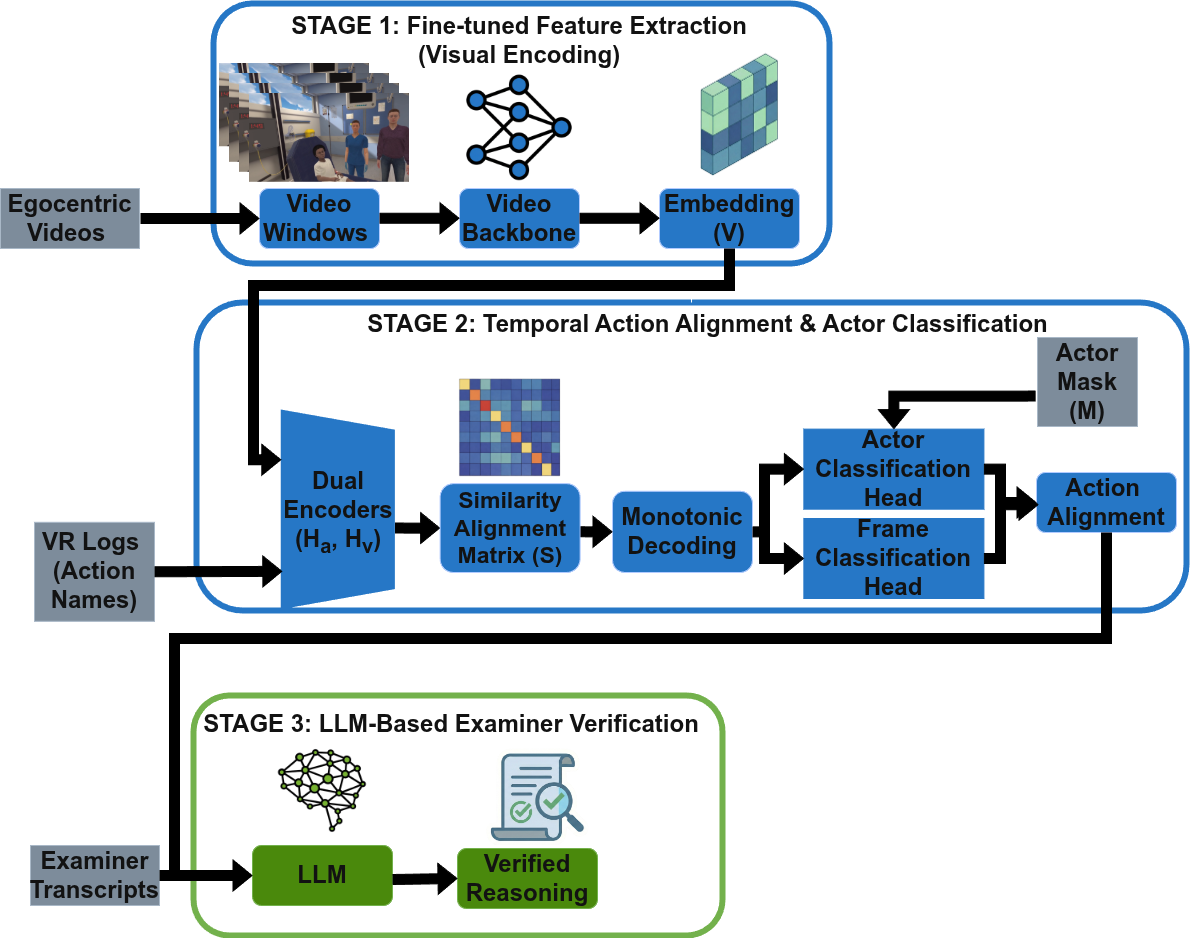}
    \caption{QAA Pipeline: Feature extraction feeds a temporal action alignment model, which produces the predicted record used for LLM verification.}
    \label{fig:fig3}
\end{figure}

The QAA framework facilitates the verification of examiner reasoning via a three-stage pipeline: (1) Fine-tuned feature extraction, (2) Temporal Action Alignment, and (3) LLM-based verification shown in Figure~\ref{fig:fig3}.

\subsection{Feature Extraction}
We fine-tuned three video backbones: X3D \cite{feichtenhofer2020x3d}, SlowFast \cite{feichtenhofer2019slowfast}, and VideoMAE \cite{tong2022videomae}, initialized from Kinetics-400 pre-training \cite{kay2017kineticshumanactionvideo} on 32-frame windows of egocentric video, slid at a stride of 8 frames so that consecutive windows overlap. Backbone feature maps are spatially pooled, projected to 768 dimensions, and temporally pooled to a single embedding per window.

A lightweight Transformer head \cite{NIPS2017_3f5ee243} (4 layers, 8 heads) processes sequences of consecutive window embeddings after each backbone, trained on two objectives: (1) Action Presence, whether a logged action starts within the window, using focal cross-entropy \cite{bib_focal_loss} ($\gamma{=}2$); (2) Actor Classification of that action over \{Child, Carer, Nurse, Item\}, also with focal cross-entropy. At evaluation, a window is attributed to an actor only when the presence head fires and to a background class otherwise. For downstream alignment we discard the heads and export the per-window 768-dimensional backbone embeddings $\mathbf{V}$, so the alignment stage receives purely visual features. Because the backbone is fine-tuned end-to-end through these heads, the exported features already encode actor identity as well as action presence.

\subsection{Temporal Action Alignment Model}
We frame alignment as a structured prediction problem that maps the ordered sequence of $N$ actions $\mathbf{A} = (\mathbf{a}_1, \dots, \mathbf{a}_N)$ recorded in an individual student's VR log onto the $T$ overlapping stride-8 video windows summarized by the extracted features $\mathbf{V} = (\mathbf{v}_1,\dots,\mathbf{v}_T)$. The log supplies only the names and order of the actions that student performed (repeated actions appear as separate entries, and actions never performed do not enter $\mathbf{A}$, so the model cannot fabricate an occurrence), but neither precise timing nor target actors; the model must therefore recover, for every logged action, (i) the window in which it occurs, (ii) the actor it is directed at, and (iii) a sub-window start offset, while guaranteeing that the recovered timeline is temporally monotonic and physically plausible.

\paragraph{Dual Encoders.}
Two Transformer encoders project the modalities into a shared $d$-dimensional space. An action encoder $E_a$ embeds each scripted action name as a learned vocabulary token and contextualizes the ordered action sequence, yielding action embeddings $\mathbf{h}_a^{(i)} \in \mathbb{R}^{d}$; a video encoder $E_v$ projects and contextualizes the window features, yielding window embeddings $\mathbf{H}_v = (\mathbf{h}_v^{(1)},\dots,\mathbf{h}_v^{(T)})$. We score every action--window pair with scaled dot-product attention,
\begin{equation}
    s_{i,t} = \frac{\langle \mathbf{h}_a^{(i)},\, \mathbf{h}_v^{(t)}\rangle}{\sqrt{d}},
\end{equation}
yielding the similarity matrix $\mathbf{S}\in\mathbb{R}^{N\times T}$, the local evidence for alignment.

\paragraph{Constrained Monotonic Alignment.}
An alignment is a path $\mathcal{P}=(t_1,\dots,t_N)$ assigning action $i$ to window $t_i$. We restrict the hypothesis space to paths that are strictly ordered and separated by at least a minimum gap $g$, i.e. $t_{i}\geq t_{i-1}+g$, preventing two distinct actions from collapsing onto the same window which is the key departure from standard Connectionist Temporal Classification (CTC) \cite{graves2006ctc}, which permits repeats and blanks. We set this gap to the minimum temporal separation between distinct actions in our data, $32$ frames; at a stride of $8$ frames this is $g{=}4$ windows. The same spacing motivates the $32$-frame feature window: no window contains more than one action, so each action-presence label is unambiguous. We score a path additively, $\mathrm{score}(\mathcal{P})=\sum_{i=1}^{N} s_{i,t_i}$, and place a Gibbs distribution $p(\mathcal{P}\mid\mathbf{S})\propto \exp(\mathrm{score}(\mathcal{P}))$ over the constrained set of paths. The log-partition function is computed in $\mathcal{O}(NT)$ by the forward recurrence
\begin{equation}
    \alpha_{i,t} = s_{i,t} + \log \sum_{t'=0}^{t-g} \exp(\alpha_{i-1, t'}),
\end{equation}
initialized with $\alpha_{1,t}=s_{1,t}$, where $\alpha_{i,t}$ accumulates the log-sum of scores over all valid partial paths that align the first $i$ actions with action $i$ placed at window $t$. The normalizer is $\log Z = \log\sum_{t}\exp(\alpha_{N,t})$. Given the reference path $\mathcal{P}^\star$, the sequence of windows containing each action's manually annotated onset, we train by minimizing the negative log-likelihood
\begin{equation}
    \mathcal{L}_{\mathrm{align}} = \log Z - \mathrm{score}(\mathcal{P}^\star),
\end{equation}
which concentrates probability mass on the annotated timeline while marginalizing all competing monotonic hypotheses.

\paragraph{Decoding.}
At inference the reference path is unavailable, so we recover the most probable alignment with a Viterbi pass under the same gap constraint: we replace the log-sum-exp in the forward recurrence with a $\max$, store back-pointers, and back-track from $\arg\max_t \alpha_{N,t}$, returning every window index $t_i$ in $\mathcal{O}(NT)$ time with a monotonic, gap-respecting timeline by construction.

\paragraph{Actor Attribution with Knowledge Constraints.}
For each aligned action we attribute the target actor. A cross-attention head queries the global video context $\mathbf{H}_v$ with the action embedding $\mathbf{h}_a^{(i)}$ and passes the result through a multilayer perceptron to produce actor logits $\mathbf{z}_i\in\mathbb{R}^{4}$ over \{Child, Carer, Nurse, Item\}. Because the scenario makes many action--actor pairings impossible (e.g., a history cannot be sourced from an Item), we inject scenario knowledge as a hard mask $\mathbf{M}$: entries of clinically impossible pairs are set to $-\infty$ before the softmax, $\hat{\mathbf{p}}_i=\mathrm{softmax}(\mathbf{z}_i+\mathbf{m}_i)$, confining all probability mass to admissible actors. This mask is what distinguishes a genuine Source Misattribution from unconstrained ambiguity. Attribution is supervised with cross-entropy $\mathcal{L}_{\mathrm{actor}}$ against the relabeled actor targets.

\paragraph{Sub-window Localization and Objective.}
A parallel head localizes the action onset within the aligned 32-frame window by classifying it into one of the $32$ candidate frame positions $\delta_i \in \{0,\dots,31\}$, decoded by $\arg\max$ at inference; it is trained with a cross-entropy loss $\mathcal{L}_{\mathrm{off}}$ against the annotated offset. The three heads are optimized jointly under the weighted objective
\begin{equation}
    \mathcal{L} = \mathcal{L}_{\mathrm{align}} + \lambda_{\mathrm{actor}}\,\mathcal{L}_{\mathrm{actor}} + \lambda_{\mathrm{off}}\,\mathcal{L}_{\mathrm{off}},
\end{equation}
so that timing, attribution, and onset are learned end-to-end over the shared dual encoders.

\subsection{LLM Examiner Verification}
The final stage treats each examiner transcript as a set of claims to be checked against the reconstructed record, casting examiner verification as grounded claim verification \cite{thorne-etal-2018-fever,min-etal-2023-factscore}. For every assessment, a large language model (LLM) receives (1) the scenario's action vocabulary, (2) the actor-attributed action record predicted by the alignment model for that student, rendered as actor--action pairs, and (3) the examiner's transcribed verbalizations from both protocol phases. A single structured prompt instructs the LLM to extract every clinical action the examiner asserts, map it onto the vocabulary, and compare it against the predicted record; it includes worked examples of a correct match, a Source Misattribution, and an Inferential error. Negative constraints filter language that does not constitute a verifiable claim: vague bundles (e.g., ``doing an A2E approach''), negations, and missed-opportunity critiques are all ignored. The LLM returns structured JSON tuples (actor, action, error type, rationale, transcript quote) under JSON-constrained decoding with default sampling. Providing the actor-attributed record is essential: actor mismatches surface as Source Misattribution, otherwise undetectable, and actions absent from the record as Inferential errors. Detections are scored against the annotated errors by greedy one-to-one matching per transcript: two items match when any of their normalized text fields, covering the action name, the transcript quote, and the stated rationale, are related by substring containment or exceed a character-level similarity of $0.8$. Overall detection credits locating the error regardless of the type the verifier assigned it; the per-type columns additionally require the annotated type. A transcript is fully corrected when every annotated error is detected.

\section{Experiments and Results}
\subsection{Experimental Setup}
All stages use student-level 5-fold cross-validation with a fixed seed of 42 across Python, NumPy, and PyTorch (including CUDA). Feature backbones were fine-tuned with AdamW \cite{kingma2017adammethodstochasticoptimization} (learning rate $5\times10^{-5}$, weight decay $10^{-4}$) for up to 50 epochs with early stopping (patience 3) on validation Actor F1, using stratified video-level folds; sequence lengths $\{5, 10, 15\}$ used batch sizes 8, 4, 2. The alignment model ($d{=}256$; a 2-layer video encoder and a 1-layer action encoder, 4 heads each) was trained with AdamW (learning rate $3\times10^{-4}$) for up to 30 epochs with early stopping (patience 3) on the product of W@16 and actor macro-F1, enforcing the 32-frame action gap with objective weights $\lambda_{\mathrm{actor}}{=}1.0$ and $\lambda_{\mathrm{off}}{=}0.5$. Leakage is prevented fold-by-fold: features for each video are extracted by the backbone that held it out as test data, and the alignment predictions used downstream come from the fold in which that student was held out, so no stage is evaluated on students it trained on. Experiments used PyTorch 2.7.1 (CUDA 12.6, cuDNN 9.0.5) on a single NVIDIA RTX 6000 Ada GPU (48\,GB) under Ubuntu 24.04 (Intel Xeon w5-2565X, 64\,GB RAM). Baselines compared each fine-tuned backbone against Kinetics-400 pre-trained weights \cite{kay2017kineticshumanactionvideo}. We report \textbf{Raw F1} (without mask), \textbf{Masked F1} (with mask $\mathbf{M}$), and \textbf{W@16}, which counts a prediction as correct if the absolute difference between predicted and annotated action start frames is $\leq 16$ frames ($\approx$0.5\,s). For transcript verification, we evaluated GPT-5.2 (XHigh Reasoning) \cite{openai2025introducingGPT52}, Kimi-k2-thinking \cite{moonshotai2025kimiK2Thinking}, and DeepSeek-v3.2 \cite{deepseekv3pushing2025}, alongside a keyword-matching baseline. For each action in the scenario vocabulary, this baseline builds lexical triggers from the action name and common clinical abbreviations, flags the action whenever a trigger occurs as a substring of the transcript, and infers the target actor from words in the surrounding context (``child''/``patient''\,$\rightarrow$\,Child, ``mum''/``carer''\,$\rightarrow$\,Carer, otherwise the vocabulary default). It treats every lexical mention as a claimed action without consulting the record, isolating how much performance requires grounded reasoning rather than surface lexical overlap. All verifiers are scored against the annotated errors under the identical matching protocol. To assess grading impact, we apply each method's corrections deterministically: every annotated grade-affecting error carries a domain and signed direction, and correction reverses that direction by one step on the three-point domain scale (net per student--examiner--domain, clipped to the scale). We then recompute the borderline-regression cut-score and test the resulting grade changes with an exact sign test over the 30 per-student composites (primary, as the 120 transcript-level scores share students and examiners) and a paired Wilcoxon signed-rank test over the 120 examiner--student checklist scores; 95\% confidence intervals for the cut-score shift come from a student-level cluster bootstrap.

\subsection{Fine-tuned Feature Extraction}
Table~\ref{tab:qa_results} reports clip-level Actor F1 and Action Recall. All backbones achieve strong actor recognition, with SlowFast performing best at sequence length 5 with an Actor F1 of $98.6\%\pm0.1$ and an action Recall of $97.6\%\pm0.3$. Gains beyond this length are limited, so we use sequence length 5 in subsequent experiments.
\begin{table*}[t]
\centering
\begin{tabular}{lccc|ccc}
\hline
 & \multicolumn{3}{c|}{Actor F1 (\%)} & \multicolumn{3}{c}{Action Recall (\%)} \\ \cline{2-7}
Model & Seq 5 & Seq 10 & Seq 15 & Seq 5 & Seq 10 & Seq 15 \\ \hline
SlowFast & $\mathbf{98.6 \pm 0.1}$ & $98.6 \pm 0.2$ & $97.9 \pm 0.3$ & $\mathbf{97.6 \pm 0.3}$ & $97.6 \pm 0.7$ & $97.0 \pm 0.4$ \\
VideoMAE & $95.4 \pm 1.4$ & $94.7 \pm 1.4$ & $94.2 \pm 1.3$ & $93.0 \pm 1.1$ & $93.0 \pm 1.4$ & $93.2 \pm 1.5$ \\
X3D & $97.3 \pm 0.2$ & $96.9 \pm 0.3$ & $94.6 \pm 3.1$ & $95.2 \pm 0.5$ & $95.1 \pm 0.8$ & $94.1 \pm 2.1$ \\ \hline
\end{tabular}
\caption{Cross-validation results for feature extractors. Best results in bold.}
\label{tab:qa_results}
\end{table*}

\subsection{Temporal Action Alignment}
Table~\ref{tab:temporal_results} summarizes temporal action alignment. The actor constraint mask increases Masked F1 by removing clinically impossible action actor pairs, but does not by itself resolve temporal ambiguity. Fine-tuning substantially improves both W@16 and Raw F1 relative to the Kinetics baseline, indicating that domain adaptation is necessary. SlowFast again achieves the strongest performance (W@16 $93.4\%\pm1.9$, Raw F1 $95.1\%\pm4.1$, Masked F1 $99.2\%\pm0.7$), and we therefore use its predictions to ground examiner verification.

\begin{table*}[t]
\centering
\begin{tabular}{lcc|cc|cc}
\hline
 & \multicolumn{2}{c|}{W@16 (\%)} & \multicolumn{2}{c|}{Raw F1 (\%)} & \multicolumn{2}{c}{Masked F1 (\%)} \\ \cline{2-7}
Model & Base & Seq 5 & Base & Seq 5 & Base & Seq 5 \\ \hline
SlowFast & $79.9 \pm 1.8$ & $\mathbf{93.4 \pm 1.9}$ & $67.3 \pm 2.9$ & $\mathbf{95.1 \pm 4.1}$ & $97.5 \pm 1.0$ & $\mathbf{99.2 \pm 0.7}$ \\
VideoMAE & $69.8 \pm 2.7$ & $69.1 \pm 7.0$ & $51.0 \pm 6.9$ & $69.1 \pm 17.9$ & $96.1 \pm 1.0$ & $95.4 \pm 3.7$ \\
X3D & $64.9 \pm 2.5$ & $93.1 \pm 1.8$ & $67.6 \pm 3.9$ & $92.0 \pm 3.0$ & $95.4 \pm 1.3$ & $98.4 \pm 0.8$ \\ \hline
\end{tabular}
\caption{Temporal Action Alignment Performance. Best results in bold.}
\label{tab:temporal_results}
\end{table*}

\begin{figure}[t!]
\centering
\includegraphics[width=0.98\columnwidth]{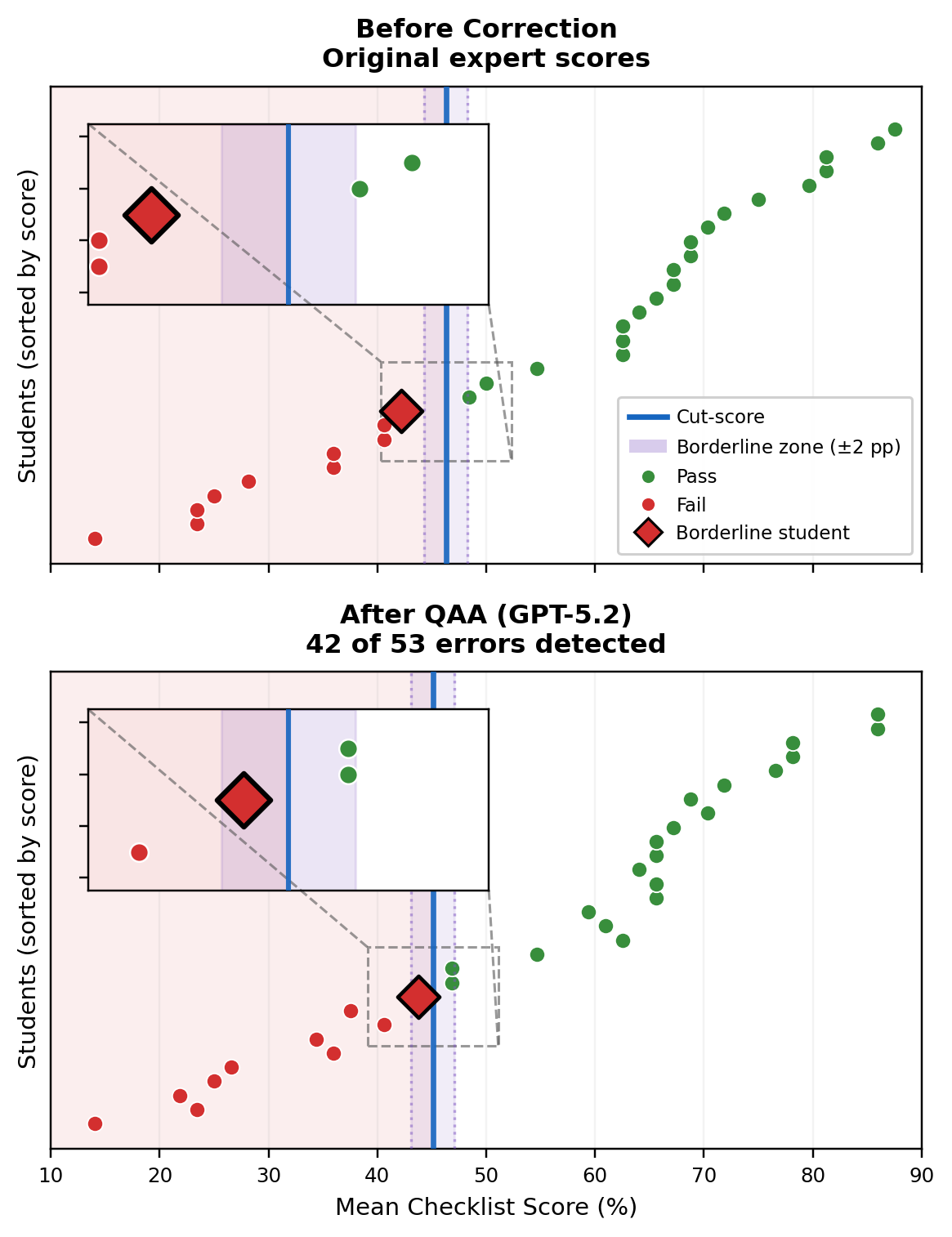}
\caption{Borderline regression before (top) and after (bottom) QAA correction. Each point is a student's mean checklist score, colored by pass/fail against the cut-score (blue line); the shaded band is the $\pm2$\,pp borderline-review zone, magnified in the insets.}
\label{fig:shift}
\end{figure}

\subsection{LLM Examiner Verification and Impact}
Using SlowFast-aligned predictions, Table~\ref{tab:results_set1} reports error detection performance. As QAA is decision-support, the LLM verifier must detect incorrect claims while avoiding excessive flags that increase reviewer burden; we therefore prioritize Inferential Recall (catching unsupported actions) and Source Misattribution Precision (avoiding incorrect actor accusations). Under this framing, GPT-5.2 offers the best balance, achieving 69.9\% precision and 76.7\% recall for overall detection, with an Inferential recall of 66.7\% and a Source Misattribution precision of 77.6\%, while producing the fewest false positives (0.44 per transcript). It fully corrects 48 of 73 error-containing transcripts, improving factual correctness from 39.2\% to 79.2\%. The gap to the alternatives is wide: the keyword baseline reaches only 2.8\% precision at 6.29 false positives per transcript, and the other LLMs either miss more errors (Kimi-k2, 39.8\% recall) or add more noise (DeepSeek, 20.5\% precision). Reliable detection therefore depends on grounded reasoning over the actor-attributed record, not surface lexical cues.

\begin{table*}[t]
\centering
\begin{tabular}{|l|cc|cc|cc|c|c|}
\hline
 & \multicolumn{2}{c|}{\begin{tabular}[c]{@{}c@{}}Overall \\ Detection\end{tabular}} & \multicolumn{2}{c|}{Inferential} & \multicolumn{2}{c|}{\begin{tabular}[c]{@{}c@{}}Source \\ Misattribution\end{tabular}} & \multirow{2}{*}{\begin{tabular}[c]{@{}c@{}}Fully \\ Corrected\end{tabular}} & \multirow{2}{*}{\begin{tabular}[c]{@{}c@{}}Average\\ FPs\end{tabular}} \\ \cline{2-7}
Approach & P (\%) & R (\%) & P (\%) & R (\%) & P (\%) & R (\%) &  &  \\ \hline
Keyword Baseline & 2.8 & 21.4 & 2.3 & 35.3 & 0.5 & 7.7 & 12 & 6.29 \\
GPT-5.2 (XHigh) & \textbf{69.9} & \textbf{76.7} & \textbf{61.8} & \textbf{66.7} & \textbf{77.6} & \textbf{86.5} & \textbf{48} & \textbf{0.44} \\
Kimi-k2-thinking & 27.2 & 39.8 & 25.9 & 27.5 & 27.8 & 51.9 & 22 & 1.21 \\
DeepSeek-v3.2 & 20.5 & 61.2 & 30.3 & 45.1 & 17.3 & 76.9 & 34 & 1.21 \\ \hline
\end{tabular}
\caption{LLM Verification Results. Best results in bold.}
\label{tab:results_set1}
\end{table*}
To quantify downstream impact, we recomputed borderline regression under each correction method (Table~\ref{tab:brm_student}, Figure~\ref{fig:shift}). The effect is broad: a full correction of all 53 errors changes the composite score of 23 of 30 students (20 fall, 3 rise; mean $-1.4$pp, up to $-4.7$pp), and the errors GPT-5.2 detects alone move 20 of 30. Examiner factual errors thus perturb most of the cohort's grades, and these shifts are statistically significant, per student (exact sign test on the 30 composites, $p<10^{-3}$) and per transcript (paired Wilcoxon over the 120 examiner--student scores, $p<10^{-4}$), whereas the keyword baseline moves only 4 students and does not change grades detectably ($p=0.10$), and the few changes it does make nudge the cut-score upward rather than down. The harm concentrates at the borderline, where the cut-score decides the outcome. Because the errors are predominantly inflations, they raise the cut-score and push marginal students down: without QAA the cut-score is 46.3\% and a borderline student sits $-4.1$pp below it, a clear fail well outside the $\pm2$pp review zone, produced by inflations in other students' assessments together with two of their own actions being wrongly downgraded. Correcting the errors GPT-5.2 verifies lowers the cut-score to 45.1\% and lifts the student to $-1.3$pp, out of an automatic fail and into the review zone where a human adjudicates; the full (oracle) correction moves them across the line entirely ($+0.1$pp). Under the annotation-based counterfactual, a pass/fail decision was changed by examiner error rather than by the student's own performance: exactly the unfairness QAA exists to surface.

\begin{table}[t]
\centering
\setlength{\tabcolsep}{1mm}
{
\begin{tabular}{l|c|c|c|c}
\hline
Method & Det. & $\Delta$cut [95\% CI] & $p$ & Student \\ \hline
No QAA & 0 & --- & --- & $-4.1$ (fail) \\
Keyword & 7 & $+0.3$ $[0.0,0.5]$ & $0.10$ & $-4.4$ (fail) \\
Kimi & 27 & $-1.0$ $[-1.5,-0.6]$ & ${<}10^{-4}$ & $-3.1$ (fail) \\
DeepSeek& 35 & $-1.1$ $[-1.6,-0.6]$ & ${<}10^{-4}$ & $-3.0$ (fail) \\
GPT-5.2 & 42 & $-1.2$ $[-1.8,-0.7]$ & ${<}10^{-4}$ & $\mathbf{-1.3}$ (rev.) \\
Oracle & 53 & $-1.1$ $[-1.8,-0.4]$ & ${<}10^{-4}$ & $+0.1$ (pass) \\ \hline
\end{tabular}
}
\caption{Borderline-regression impact and significance of the grade shift. $\Delta$cut is the cohort cut-score change (pp), with a 95\% interval from a student-level cluster bootstrap; $p$ is a paired Wilcoxon signed-rank test over the 120 examiner--student checklist scores (corrected vs.\ original). ``rev.''\ = borderline-review zone. Det.\ is the number of grade-affecting errors corrected; Kimi, DeepSeek, and GPT-5.2 abbreviate Kimi-k2-thinking, DeepSeek-v3.2, and GPT-5.2 (XHigh).}
\label{tab:brm_student}
\end{table}

\section{Discussion and Conclusion}
We introduced Quality Action Assurance (QAA), a multimodal framework that shifts the analytical target in medical assessment from students to the verification of examiners, grounding examiner verbalizations against an actor-attributed event record. Our cognitive analysis reveals a reliability paradox: examiners exhibit substantial-to-near-perfect agreement (AC$_2$ $\approx$ 0.65--0.86) yet make factual errors in over 60\% of examinations, including grade-affecting errors across most domains. Standard safeguards miss this entirely, because inter-rater agreement measures whether examiners are \emph{consistent}, not whether they are \emph{correct}.

These errors are not rare slips affecting a few students: correcting them moves the grades of more than three-quarters of the cohort. The harm, however, falls hardest at the decision boundary, where a one- or two-point shift is immaterial for a clear pass or fail but decides a borderline outcome, as our borderline case illustrates. The significance tests confirm these shifts are real rather than noise, yet the substantive point is fairness at the margin, not effect size.

This vulnerability is unlikely to be specific to our scenario. Borderline regression from domain scores to a global outcome is the standard-setting method for OSCEs, and the same approach is used in surgical and other observational examinations; any assessment that aggregates human judgments into a cut-score inherits the same failure mode, in which systematic examiner error silently shifts the boundary. Because QAA depends only on synchronized video and action logs rather than on the menu-driven interface of our study, it is potentially applicable wherever such a record can be reconstructed; transfer to voice-based and less structured examinations remains untested.

We position QAA as decision-support, not automation. It does not grade students or overrule examiners; it converts verbalized reasoning into time-stamped, actor-attributed claims and surfaces the specific discrepancies a human should review. This makes retrospective quality assurance tractable, narrowing attention to concrete, checkable items rather than an exhaustive audit, while guarding against automation bias: the verifier can itself be incorrect, and the record it checks against is a model prediction, so its outputs must remain advisory, calibrated to a low false-positive rate, and validated per scenario. Examiners may accept flags that confirm their view and dismiss those that do not \cite{Dietvorst2014-xp,Schemmer_2023}, which matters because QAA exists to contradict them. Where algorithmic auditing usually holds automated systems to account \cite{10.1145/3351095.3372873}, QAA uses a model to audit human decisions. Deployment therefore raises a governance question specific to examiner verification: QAA produces an attributable, time-stamped record of individual examiner factual errors. Such records are suited to aggregate quality assurance and examiner development, but should not be repurposed for individual performance management without examiner consent and institutional oversight.

Our study has limitations. The cognitive-error analysis draws on 30 students, four examiners, and 120 transcripts from a single menu-driven pediatric scenario on one platform (alignment models trained on 91 students), so error-prevalence rates and the single observed pass/fail flip are indicative rather than definitive; establishing how common such errors are requires larger multi-scenario, multi-site studies. The menu-driven modality, chosen because it yields verifiable clicks, is less representative of emerging voice-based VR OSCEs, which would require speech-to-action grounding. Error annotations came from a single researcher; because the core determination, whether a claimed action or actor appears in the objective VR log, is a factual check rather than a quality judgment, exposure to annotator bias is limited, and multi-rater validation of the more interpretive grade-impact labels is planned. The error annotations and QAA's verification target also derive from the same log and annotation process, so our evaluation measures recovery of the annotated errors rather than agreement with an independent standard. Finally, the verifier recovers 42 of 53 grade-affecting errors and checks against a predicted rather than certified record, so its outputs remain advisory and full correction requires human adjudication. Even so, QAA demonstrates that multimodal grounding makes examiner reasoning auditable, converting claims into verifiable actor--action tuples anchored to a time-aligned record; by improving factual correctness and reducing manual checking, it supports fairer standard setting across observational examinations vulnerable to cognitive bias, while preserving examiner authority through human adjudication.

\section*{Acknowledgments}
H.R. and A.N. acknowledge the EPSRC Turing AI Fellowship ``Ultrasound
Multi-Modal Video-based Human--Machine Collaboration'' [EP/X040186/1]. We thank the examiners and medical students who participated in the VR OSCE
marking studies. The authors declare no competing interests.

\bibliography{aaai2027}

\end{document}